%% file: main.tex
\documentclass{article}

\PassOptionsToPackage{numbers, compress}{natbib}

\usepackage[preprint]{neurips_2024}

\usepackage[utf8]{inputenc} 
\usepackage[T1]{fontenc}    
\usepackage{hyperref}       
\usepackage{url}            
\usepackage{booktabs}       
\usepackage{amsfonts}       
\usepackage{nicefrac}       
\usepackage{microtype}      
\usepackage{graphicx}

\usepackage{multicol}
\usepackage{multirow}
\usepackage{csquotes}
\usepackage[table]{xcolor}
\usepackage{booktabs}
\definecolor{lightred}{RGB}{255,200,200}

\usepackage{microtype}
\usepackage{subfigure}
\usepackage{booktabs} 

\usepackage{amsmath}
\usepackage{amssymb}
\usepackage{mathtools}
\usepackage{amsthm}
\usepackage{setspace}

\title{A Dataset and Benchmark for Copyright Infringement Unlearning from Text-to-Image Diffusion Models}

\author{
  Rui Ma\thanks{Equal contribution. Corresponding to zhendong@berkeley.edu, and shanghang@pku.edu.cn}\\
  Peking University\\
  \And 
   Qiang Zhou\footnotemark[1]  \\
  Tsinghua University \\
  \And
    Yizhu Jin\\
  Beihang University\\
  \And
   Daquan Zhou\\
  Bytedance \\
  \And
    Bangjun Xiao\\
  Peking University \\
  \And
   Xiuyu Li\\
  UC Berkeley \\
   \And
   Yi Qu\\
   Edinburgh College  \\
    \And
   Aishani Singh\\
  UC Berkeley \\
  \And
   Kurt Keutzer\\
  UC Berkeley \\
    \And
    Jingtong Hu\\
  University of Pittsburgh  \\
  \And
   Xiaodong Xie\\
  Peking University  \\
   \And
   Zhen Dong\\
  UC Berkeley  \\
  \And
   Shanghang Zhang\\
  Peking University \\
    \And
   Shiji Zhou\\
   Tsinghua University\\
}

\begin{document}

\maketitle

\input{sec/0_abstract}    
\input{sec/1_intro}
\input{sec/2_formatting}
\input{sec/4_dataset}

\input{sec/5_benchmark}
\input{sec/7_conclusion}

\bibliographystyle{plain}
\bibliography{main}

\input{sec/checklist}
\input{sec/datasheets}

\end{document}

%% file: sec/0_abstract.tex
\begin{abstract}

Copyright law confers upon creators the exclusive rights to reproduce, distribute, and monetize their creative works. However, recent progress in text-to-image generation has introduced formidable challenges to copyright enforcement. These technologies enable the unauthorized learning and replication of copyrighted content, artistic creations, and likenesses, leading to the proliferation of unregulated content. Notably, models like stable diffusion, which excel in text-to-image synthesis, heighten the risk of copyright infringement and unauthorized distribution.Machine unlearning, which seeks to eradicate the influence of specific data or concepts from machine learning models, emerges as a promising solution by eliminating the \enquote{copyright memories} ingrained in diffusion models. Yet, the absence of comprehensive large-scale datasets and standardized benchmarks for evaluating the efficacy of unlearning techniques in the copyright protection scenarios impedes the development of more effective unlearning methods. To address this gap, we introduce a novel pipeline that harmonizes CLIP, ChatGPT, and diffusion models to curate a dataset. This dataset encompasses anchor images, associated prompts, and images synthesized by text-to-image models. 
Additionally, we have developed a mixed metric based on semantic and style information, validated through both human and artist assessments, to gauge the effectiveness of unlearning approaches. 
Our dataset, benchmark library, and evaluation metrics will be made publicly available to foster future research and practical applications (\href{https://rmpku.github.io/CPDM-page/}{website} / \href{http://149.104.22.83/unlearning.tar.gz}{dataset}).

\end{abstract}

%% file: sec/1_intro.tex
\section{Introduction}
\label{sec:intro}


\begin{figure}[t]
    \centering
    \includegraphics[width=0.8\linewidth]{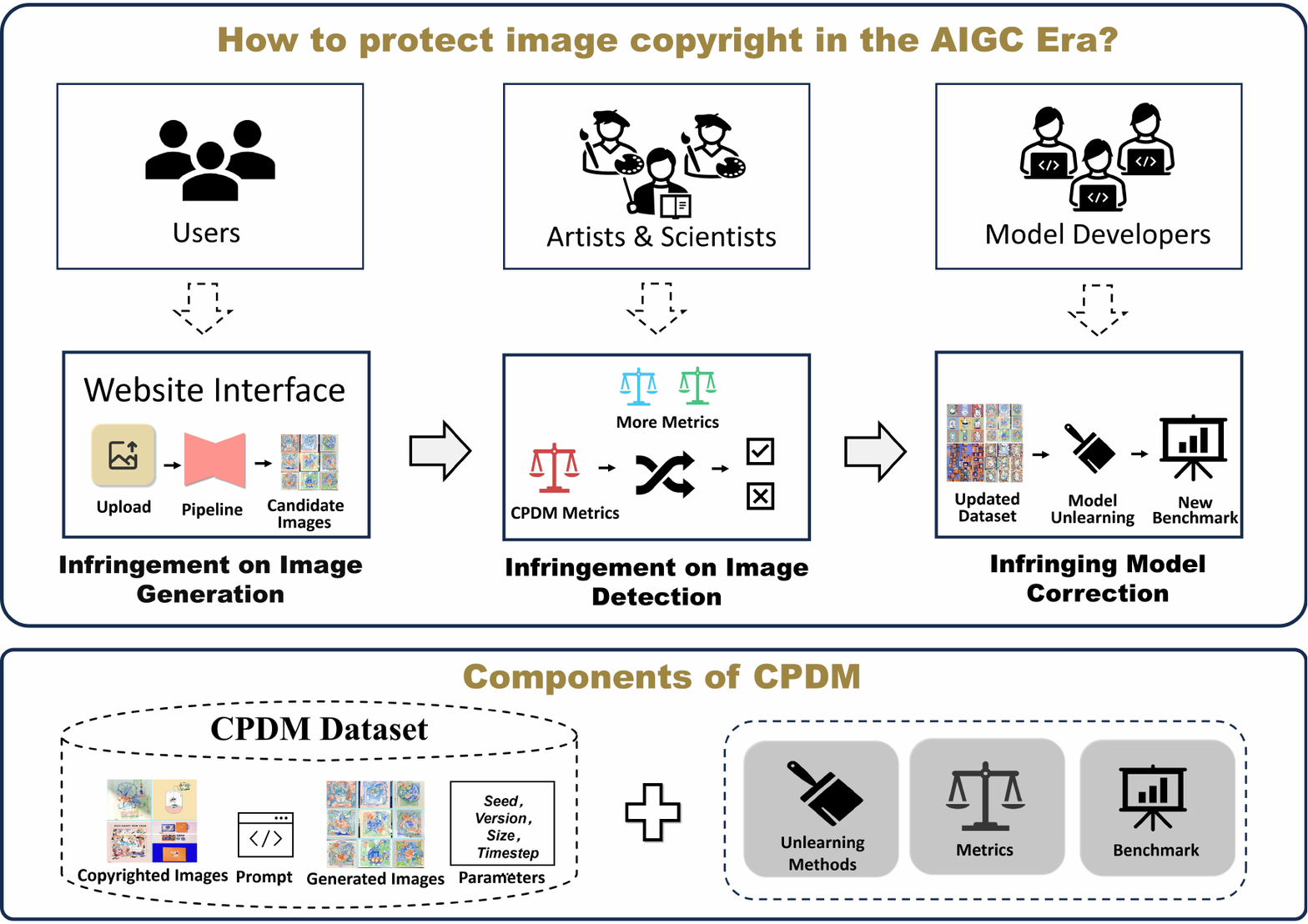}
    \caption{The pipeline of Copyright Unlearning and the crucial role of Datasets and Benchmarks.
    }
    \vspace{-0.5cm}
    \label{fig:how_to_protect}
\end{figure}

Text-to-image generative models have recently emerged as a significant topic in computer vision, demonstrating remarkable results in the area of generative modeling~\cite{goodfellow2020generative, stablediffusion}. These models bridge the gap between language and visual contents by generating realistic images from textual descriptions. However, rapid advancements in text-to-image generation techniques have raised concerns about copyright protection, particularly unauthorized reproduction of content, artistic creations, and portraits~\cite{carlini2023extracting}. A specific concern arises from the use of Stable Diffusion (SD), a state-of-the-art text-conditional latent diffusion model, which has sparked global discussions on copyright. 


Machine Unlearning (MU), which aims to eliminate the influence of specific target data or concepts, presents a promising solution to the aforementioned challenges. Several studies have provided intuitive evidence of MU's effectiveness in erasing the memory of copyrighted material from original models~\cite{zhang2023forget,gandikota2023erasing,kumari2023ablating,gandikota2024unified}. However, the current body of research is limited by a lack of quantitative and systematic assessments that evaluate the extent to which MU can reduce the risk of copyright infringement. This limitation hinders the ability to make meaningful comparisons between existing MU approaches. The challenge is exacerbated by the inherent complexity of defining copyright infringement criteria for text-to-image generative models, as well as the scarcity of comprehensive inference datasets and standardized benchmarks for assessing copyright infringement. The absence of extensive copyright datasets obstructs researchers' efforts to fully comprehend the copyright infringement risks associated with generative models. This, in turn, restricts their capacity to develop superior MU algorithms capable of addressing the legal risks effectively.
As shown in Fig. \ref{fig:how_to_protect}, datasets and benchmarks are essential for forgetting copyrighted content and evaluating unlearning methods.

Initially, it is crucial to define what constitutes copyright infringement in contents produced by text-to-image generative models~\cite{somepalli2023diffusion}. In this study, we focus on infringement within 2D artistic works. Drawing on expertise from copyright protection specialists, including artists and lawyers, we contend that a unique painting style of an artist, virtual representations in artistic creations, and individual portraits all represent forms of creative expression deserving of legal protection. In order to identify instances of infringement in these contexts, a comprehensive analysis encompassing both technical and semantic aspects of the generated contents is indispensable.

\begin{figure*}[t]
  \centering
  \includegraphics[width=0.8\linewidth]{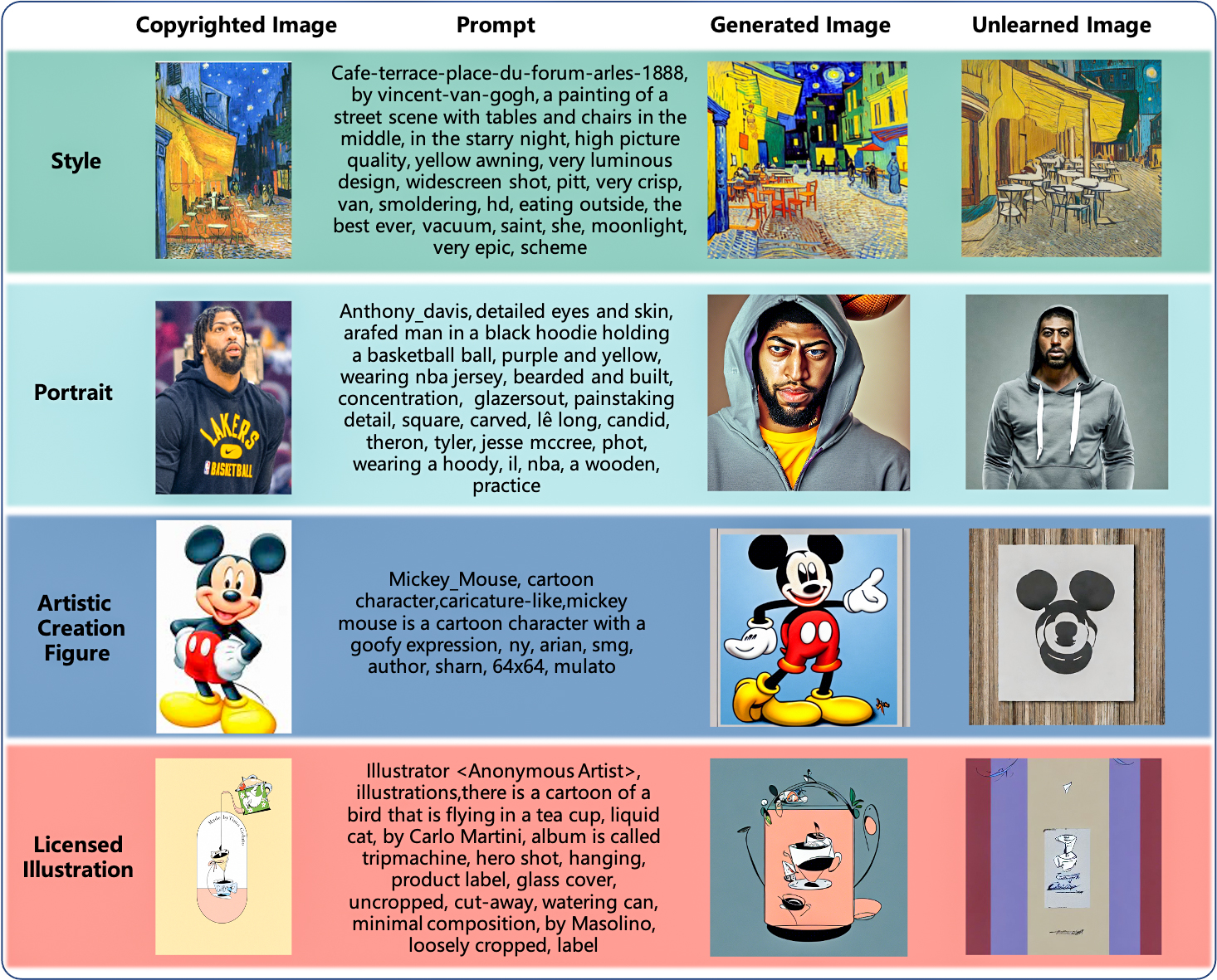}
  \caption{Examples of CPDM dataset composition and unlearned results for copyright protection. 
  }
  \vspace{-0.5cm}
  \label{fig:dataset_example}
\end{figure*}

In light of the aforementioned considerations, we introduce the first comprehensive dataset tailored for copyright protection in this domain: \textbf{C}opyright \textbf{P}rotection from \textbf{D}iffusion \textbf{M}odel (\textbf{CPDM}) dataset. Specifically, we have curated copyright images from four distinct categories that are most commonly suspected of infringement, as demonstrated in Fig. \ref{fig:dataset_example}.  
This dataset encompasses a collection of original copyright images, associated prompts for text-to-image generation via Stable Diffusion, and indicatives for a series of features, represented by: 
1) Potential for Infringement; 
2) Effectiveness of Unlearning; 
3) Extent of Model Degradation during Unlearning.

\textbf{Potential for Infringement.} This is quantified by feature-level similarities between the original copyright images and potentially infringing generated contents, denoted as CPDM metric (\textbf{CM}). 

\textbf{Effectiveness of Unlearning.} We reuse the CM metric to indicate the similarities between the original copyright images and their unlearned counterparts after processed by unlearning methods. Additionally, we evaluate changes of CLIP scores, denoted as \textbf{$\Delta$CLIP}, for text-image similarity. This indicates the extent to which the prompt that generates potential infringement is nullified. 

\textbf{Extent of Model Degradation during Unlearning.} The unlearning process inherently degrades the model by eliminating certain infringement-suspected concepts. Nevertheless, it is vital to preserve the Stable Diffusion model's generation capacities for copyright-irrelevant contents. We assess the degree of model degradation using the widely-recognized FID (Fréchet Inception Distance) metric~\cite{heusel2018gans}.

Our benchmark facilitates a straightforward evaluation of potential copyright infringement, while facilitating comparison among various unlearning methods. 
Moreover, we perform comprehensive benchmark tests on our proposed CPDM dataset. In our experiments, we utilize gradient ascent-based and response-based pruning methods for unlearning, as comparison baselines for other unlearning approaches, specifically targeting the Stable Diffusion models. 
This evaluation provides valuable insights into assessing copyright infringement and the efficacy of unlearning methods in reducing infringement risks, while preserving the ability to generate non-infringing contents.

%% file: sec/2_formatting.tex
\section{Background and Related Work}

\textbf{Text-to-image Generative Model.} 
Recently text-to-image diffusion models~\cite{stablediffusion} have emerged as a crucial research area attracting wide attention. 
These state-of-the-art methods~\cite{nichol2022glide, stablediffusion, saharia2022photorealistic, ramesh2022hierarchical,balaji2023ediffi} have exhibited remarkable capabilities in transforming textual information into visually coherent and realistic images, often demonstrating high performance in terms of accuracy. 
The advancements in these techniques have opened up a plethora of possibilities for a wide range of downstream tasks, such as image editing~\cite{kawar2023imagic, goel2023pairdiffusion, couairon2023diffedit, Avrahami_2022}, image denoising~\cite{ho2020denoising, xie2023diffusion} and super-resolution~\cite{li2021srdiff, gao2023implicit}, etc.
The advancements in text-to-image generative models have significantly impacted various industries. 
This convergence raises critical questions about authorship, intellectual property rights, and the implications for plagiarism in the digital era. Addressing these concerns is imperative.

\textbf{Image Similarity Measurement.} 
Determining plagiarism, a crucial concern in arts and legal domains, often relies on assessing image similarity. Existing metrics like PSNR and SSIM are limited to near-identical images, lacking the capacity to evaluate higher-level similarities~\cite{1284395}. Perceptual metrics like LPIPS, while aligning closely with human perception, may have limitations in capturing certain nuances\cite{zhang2018unreasonable}. In the realm of videos, metrics like VMAF, despite combining human visual modeling with machine learning techniques, may encounter challenges in specific scenarios\cite{sheikh2006image}. While FID is prevalent in generative tasks, it primarily evaluates the distance between sets of images, leaving a gap in measuring inherent similarity between anchor and generated images~\cite{heusel2017gans}. This underscores the need for further investigation into copyright protection and similarity definition.

\textbf{Unlearning Method.} 
Carlini et al. ~\cite{carlini2023extracting} highlights that the privacy of Stable Diffusion models is significantly lower compared to Generative Adversarial Networks (GANs) ~\cite{goodfellow2020generative}. 
Under the diffusion framework, models tend to retain certain images from the training data, potentially generating outputs that closely resemble the original images. 
To remove explicit artworks from diffusion models, Andikota et al.~\cite{gandikota2023erasing} presents a fine-tuning method for concept removal from diffusion models. 
Additionally, Zhang et al.~\cite{zhang2023forget} presents the \enquote{Forget-Me-Not} method, which enables the targeted removal of specific objects and content from diffusion models within a span of 30 seconds, while minimizing the impact on other contents.
Somepalli et al.~\cite{somepalli2023diffusion} explores whether diffusion models create unique artworks or directly replicate certain contents from the training dataset during image generation. 
Furthermore, there exist numerous model unlearning methods in the context of image-related tasks, as evidenced by ~\cite{bourtoule2021machine, ginart2019making, guo2019certified, graves2021amnesiac, huang2021unlearnable,kumari2023ablating,gandikota2024unified}, among others.

\textbf{Works in Artistic Image Communities.}
With the emergence of painting capabilities in models like Stable Diffusion, there has been a growing surge in activity and attention within communities focused on image copyright protection. For instance, websites such as 
\href{https://stablediffusion.fr/artists}{stablediffusion.fr/artists} and \href{https://www.urania.ai/top-sd-artists}{urania.ai/top-sd-artists} 
have gained prominence. 
In comparison to the efforts of these image communities, our approach involves the collection of real, valuable, and specific images from the art world to be used as training examples for image generation models. This represents a more rigorous form of style imitation. In contrast, the artistic images in the provided links tend to focus on capturing certain aspects of an artist's style, fitting into a broader category of style imitation. For details, see the supplementary materials.

\textbf{Policy, Legal, and Social Impact.}
The increasing global popularity of AIGC highlights the importance of privacy and copyright issues. AI companies, including OpenAI, have taken measures to address concerns related to data security. The US has proposed establishing a new government agency responsible for approving large-scale AI models. Furthermore, the Chinese Cyberspace Administration has published a document emphasizing AIGC security issues\footnote{\url{http://www.cac.gov.cn/2023-04/11/c_1682854275475410.htm}}.
Recently enacted legislation, such as the General Data Protection Regulation 
(GDPR\footnote{\url{https://gdpr-info.eu/}})
in the European Union, the California Consumer Privacy Act 
(CCPA\footnote{\url{https://oag.ca.gov/privacy/ccpa}})
in California, and the Personal Information Protection and Electronic Documents Act (PIPEDA\footnote{\url{https://laws-lois.justice.gc.ca/ENG/ACTS/P-8.6/index.html}})
in Canada, have legally solidified this right \cite{zhang2023review, chen2021machine}.

%% file: sec/4_dataset.tex
\section{CPDM Dataset}

\subsection{Dataset Creation Pipeline}
\label{pipeline for generating datasets}

\begin{figure*}[]
  \centering
  \includegraphics[width=0.9\linewidth]{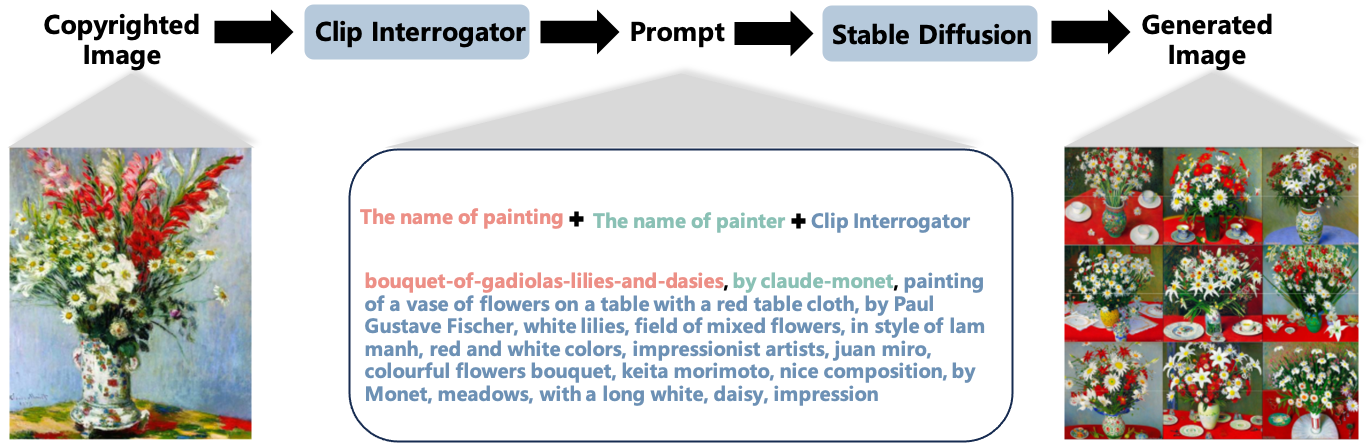}
  \caption{Pipeline for the CPDM Dataset Creation. The \textit{clip-interrogator} is utilized to convert copyrighted images into corresponding textual information. This text is subsequently refined and transformed into prompts, which are then inputted into a diffusion model to generate the corresponding infringing images.}
  \label{fig:prompt_pipeline}
  \vspace{-5mm}
\end{figure*}

We propose a pipeline to coordinate CLIP and diffusion models to generate a dataset that contains anchor images, corresponding prompts, and images generated by text-to-image models, reflecting the potential abuses of copyright, as illustrated in Fig. \ref{fig:prompt_pipeline}. 
Initially, we collect a set of images that potentially contain copyrighted content, which serves as anchor images. Subsequently, these images are fed into the \textit{clip-interrogator}\footnote{\url{https://github.com/pharmapsychotic/clip-interrogator}}, allowing us to obtain prompts that correspond to each anchor image. 
We also provided experimental results using tools other than clip-interrogator in the supplementary materials.
Finally, the prompts are used as inputs for the stable diffusion model, resulting in the generation of images by the stable diffusion model. The final outcomes indicate that even such a rudimentary pipeline can effectively generate a substantial volume of works pertinent to infringement issues. 

\begin{table*}[h]
  \vspace{-9mm}
  \caption{Statistics and Details of the CPDM dataset.}
  \label{tab:data_source}
    \setlength{\tabcolsep}{1mm}
  \centering
  \begin{tabular}{cccc}
    \toprule
    
    
    \textbf{Name}     & \textbf{Source} &
    \textbf{Num. of anchor image} & \textbf{Num. of generation} \\
    
    \midrule
    Style & From WikiArt  & $\sim$1500  & $\sim$13500  \\
    Portrait     & From Wikipedia & 200  & 1800   \\
    Artistic Creation Figure     & From Wikipedia      & 200 &1800  \\
    Licensed Illustration     & From Anonymous Artist      & 200  &1800\\
    \bottomrule
    \centering
  \end{tabular}
  \vspace{-3mm}
\end{table*}

\begin{figure*}[htbp]
  \centering
  \includegraphics[width=1\linewidth]{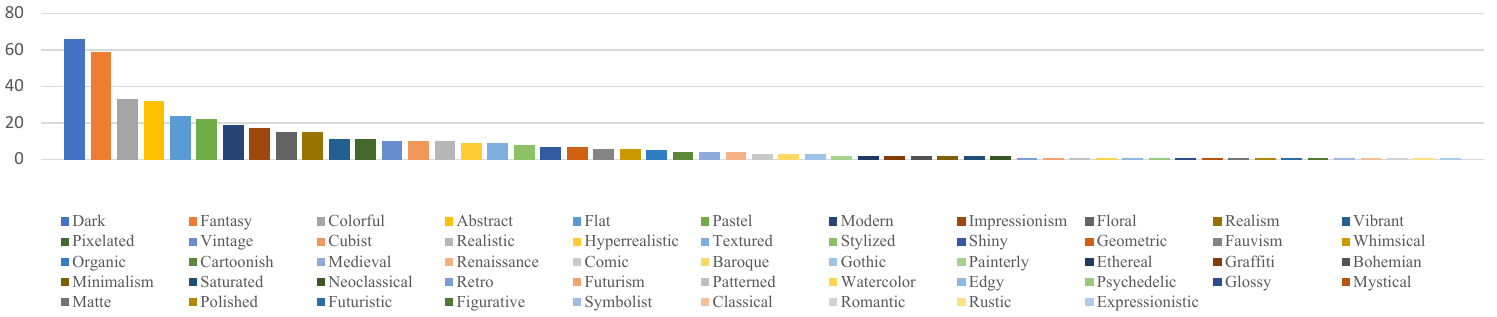}
  \caption{Analysis of Dataset Style Diversity. We conducted a statistical of the styles in prompts obtained using the pipeline in Fig. \ref{fig:prompt_pipeline}. The various styles at the bottom are provided by artists.}
  \label{fig:cpdm-dataset-static}
  \vspace{-3mm}
\end{figure*}

\subsection{Dataset Composition}

After engaging in thorough discussions with both artists and legal experts, we divided the dataset into the following four parts, to facilitate a comprehensive examination from various viewpoints to identify and address potential copyright infringement concerns effectively. The sources of our curated data are listed in Tab. \ref{tab:data_source}.
The style diversity statistics of the dataset are shown in Fig. \ref{fig:cpdm-dataset-static}.

\textbf{Style. }
Painting artworks often embody the distinctive style of the artist, encompassing aspects such as brushstrokes, lines, colors, and compositions. This artistic style is also a form of copyright that requires protection. WikiArt is an online user-editable visual art encyclopedia, a source from which numerous art-related datasets~\cite{karayev2014, liao2022artbench} have been curated. WikiArt already features some 250,000 artworks by 3,000 artists, localized on 8 languages. These artworks are in museums, universities, town halls, and other civic buildings of more than 100 countries. We selected approximately 1500 artworks from 100 artists on WikiArt as the source of anchor images for prompt generation and corresponding content generation using stable diffusion.

\textbf{Portrait. }
The right of portrait refers to an individual's control and use of their own portrait, including facial features, image, and posture. The purpose of publicity rights is to safeguard an individual's privacy, personal dignity, and image integrity, preventing unauthorized use, disclosure, or alteration of their portrait. This legal protection aims to preserve an individual's control over how their likeness is portrayed and commercialized. We utilized web scraping techniques to collect over 200 portrait images from Wikipedia, which is a free, web-based, multilingual encyclopedia that contains articles on a wide range of topics.

\textbf{Artistic Creation Figure. }
Artistic creations, including characters from animations and cartoons, are often protected by law. In this context, we refer to this category as \enquote{artistic creation figures}. Similar to portraits, we have curated a dataset of 200 influential animated characters and figures by collecting information from reputable sources such as Wikipedia.

\textbf{Licensed Illustration.}
We have obtained authorization to use a portion of Anonymous Artist's artworks in this study. Therefore, we can include these artworks as part of the training dataset for fine-tuning stable diffusion, which will be utilized for simulating infringing artistic paintings.

%% file: sec/5_benchmark.tex
\section{CPDM Benchmark}
\label{CPDM Benchmark}

\input{sec/3_finalcopy}

\subsection{Benchmark Unlearning Methods}

To facilitate fair comparison among existing unlearning methods using our proposed CPDM benchmark, we present two intuitive baseline unlearning methods: a gradient ascent-based approach and a weight pruning-based approach.

\textbf{Gradient Ascent-based Approach (\textit{GA}). } 
For a single image, forgetting can be achieved by optimizing for a few epochs with an appropriate learning rate~\cite{golatkar2020eternal,jang2023knowledge}. More specifically, for a diffusion model with its set of weight parameters $\theta$, to forget the image $Y$ and its corresponding prompt $X$,  we update $\theta$ each epoch in the following way:
\[
    \theta = \theta + \eta \nabla_{\theta} L_{mse}(\theta, X, Y)
\]
where $\eta$ is the learning rate and $L_{mse}(\theta, X, Y)$ refers to the loss computed between the generated output using prompt $X$ and the targeted image $Y$. 
It is evident that the use of gradient ascent optimization has a certain impact on the generative model's capability, even though we only optimize for a small number of epochs.

\textbf{Weight Pruning-based Approach (\textit{WP}). }
Inspired by magnitude pruning~\cite{Han2015DeepCC, han2015learning}, the core idea behind parameter pruning for forgetting certain infringing images is to mask the weights in the model so that those weights exhibit the strongest response in generating those images.
Initially, the image is passed through Stable Diffusion for forward propagation, simultaneously obtaining gradients for each network layer. Each layer is treated as a separate pruning group. The highest $p_c\%$ activation values within each layer are identified, and weights correlated with these values are located. Subsequently, based on the gradient magnitudes of these weights, the top $p_w\%$ are set to zero.
This process can be described by the following equation:
\[
    \theta^* = optim\{\theta|p_{c}*W_{ij}, \nabla_{W} L_{ij}, p_{c}*|Y_{ij}|\}
\]
To illustrate, for a layer expressed as $Y = WX$ where $W$ represents the weight (bias term omitted for simplicity), we first select $W_i$ corresponding to the greatest $p_c\%$ of $|Y_{ij}|$, where $|\cdot|$ represents the absolute value operator, $optim\{\cdot\}$ represents updating parameters. Then, for each $W_i$, we prune $p_w\%$ of the elements $W_{ij}$ corresponding to the highest $m\%$ of its gradient values $\nabla_{W} L_{ij}$, setting these to zero.
More related research work and experiments can be found in the supplementary materials.

\section{CPDM Experiments}

\subsection{Experiment Setting}

Our experiments can be roughly categorized into two perspectives. \textbf{First}, indicating the potential of infringement of SD-Generated images, using the original version of SD (SD-\href{https://huggingface.co/CompVis/stable-diffusion-v1-4}{v1.4}, SD-\href{https://github.com/CompVis/stable-diffusion}{v2.1}) and the finetuned counterpart (SD-Finetuned) respectively. Note that we directly test the SD generative models using the data from WikiArt and Wikipedia, as shown in Tab. \ref{benchmark_performance}' top. We test both the SD-Generated and SD-Finetuned models on the Illustration part of our CPDM dataset, from an anonymous artist, as shown in Tab. \ref{benchmark_performance}' bottom. 
\textbf{Second}, we conduct a comprehensive evaluation of unlearning methods, regarding their ability to remove copyright-relevant concepts and maintain the copyright-irrelevant generative capacities. In this stage, we incorporate four representative unlearning methods tailored for text-to-image generative tasks, ESD\cite{gandikota2023erasing}, FMN\cite{zhang2023forget}, CA\cite{kumari2023conceptablation}, and UCE\cite{gandikota2024unified} (further details on the setups are provided in the supplementary materials), along with our proposed baseline unlearning methods, GA and WP. 
As shown in Tab \ref{benchmark_performance}, the CPDM metrics are evaluated on a case-by-case basis, comparing anchored copyright images to the generated images produced by SD models and their unlearned counterparts, forming image pairs of <Image$_{cpr}$, Image$_{gen}$> and <Image$_{cpr}$, Image$_{unl}$>, respectively. Specifically, FID is calculated by generating 10,000 images using each model on the COCO-10K dataset \cite{lin2014microsoft}. 
In the context of an unlearning process, for our proposed two baseline unlearning methods, we only make parameter adjustments on the U-Net structure of SD models\cite{rombach2022high}, while freezing the other parameters.

\subsection{Benchmark Metric Evaluation}
\label{sec:effectiveness}

\begin{figure*}[t]
  \centering
   \includegraphics[width=1\linewidth]{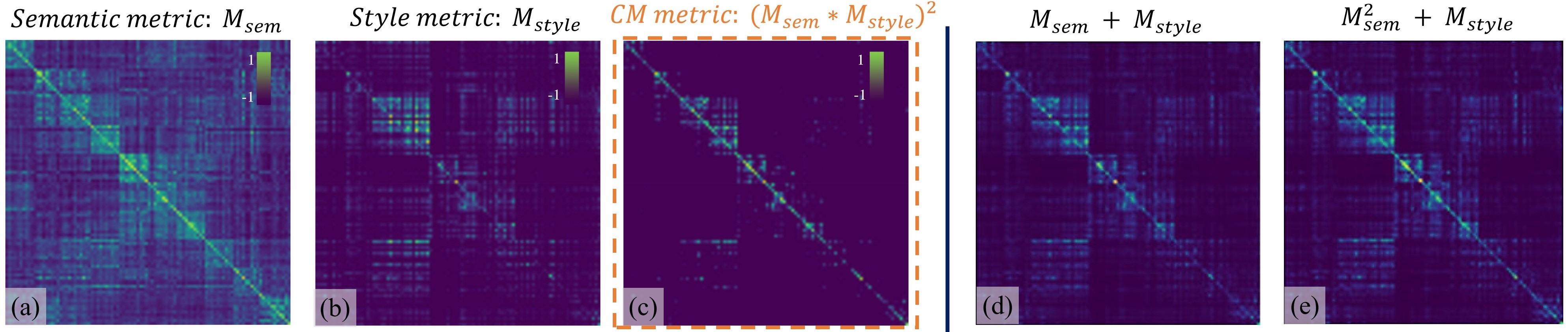}
  \caption{
  Analysis and Comparison of the Effectiveness of Evaluation Metrics.
  }
  \vspace{-5mm}
  \label{fig:metric_analysis}
\end{figure*}

In Fig. \ref{fig:metric_analysis}, we selected ten artists and generated images that mimic their artwork styles using the pipeline described in Fig. \ref{fig:prompt_pipeline}. For each artist, ten generated images were randomly chosen, resulting in a total of 100 images. 
We then used our metrics to compute a matrix ($M^{100 \times 100}$), capturing the relationships between the generated images and their corresponding anchor images.
The horizontal axis representing anchor images and the vertical axis representing generated images. Brighter pixels indicate higher similarity between image pairs. A brighter diagonal suggests a more accurate metric assessment.
In (c), the highlighted pixels primarily near the diagonal confirm the CM metric's effectiveness in assessing image similarity. Additionally, (a) and (b) illustrate the Semantic metric's insensitivity to certain dissimilarities, while the Style metric is overly sensitive to some variations. 
The results of (c) and (d) also demonstrate the rationality of the design of the style metric in Section 4.1.
These visualizations support the assertion that our metric can identify images that may constitute plagiarism. Moreover, this approach can be applied to unlearning tasks, providing a systematic and quantifiable method for evaluating unlearning techniques.

\begin{figure*}[h]
  \centering
  \includegraphics[width=1\linewidth]{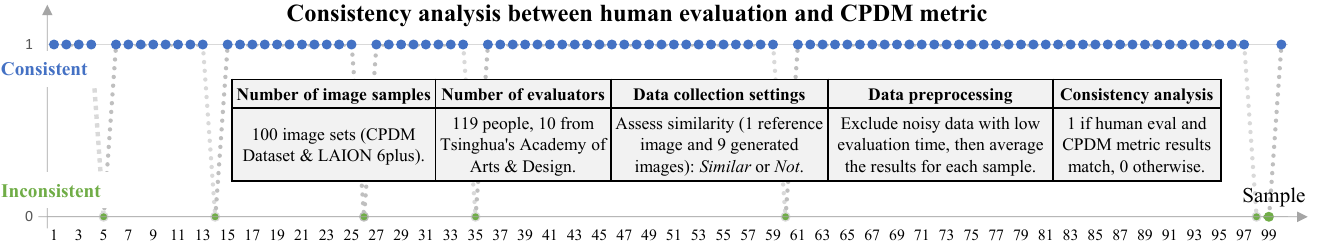}
  \caption{Consistency analysis between human evaluation and CM metric.}
  \label{fig:consistency_analysis}
  \vspace{-4mm}
\end{figure*}

\subsection{Human Verification}

To ensure a high-quality dataset and the design of a more compelling evaluation metric, we manually verified the dataset obtained from the creation pipeline and assessed the consistency between CM and manual evaluation. 
\textbf{1) For the CPDM dataset}, as the current dataset is relatively small and manageable for manual screening, we assembled a team of four individuals with expertise in image-related work to verify the dataset. These experts meticulously reviewed the dataset, removing dissimilar images. To ensure a high level of similarity among the remaining images, a final vote was conducted to decide the inclusion of each image in the dataset.
\textbf{2) For the CM metric}, as shown in Fig. \ref{fig:consistency_analysis}, we analyzed 100 sets of images with 119 manual evaluation results. 
The details of the experimental setup can be seen in the gray table in the image.
The experimental results indicate that our metrics achieve results fairly consistent with manual evaluations in determining image similarity.

\begin{table*}[t]
  \centering
  \caption{
   \textit{Top}: benchmark performance on WikiArt and Wikipedia.
    \textit{Bottom}: benchmark performance on CPDM Illustration.
    On the right, \textit{WP} and others represent different unlearning methods.
  }
    \setlength{\tabcolsep}{0.7mm}
  \begin{tabular}{c|c|cc|cccccc}

    \hline\toprule[1pt]

    &\multirow{2}{*}{\textbf{Metric}}& \textbf{SD-Generated} & \textbf{SD-Finetuned} & \textbf{\textit{WP}} & \textbf{\textit{GA}} & \textbf{\textit{ESD}} & \textbf{\textit{FMN}} & \textbf{\textit{CA}} & \textbf{\textit{UCE}}\\

    & &\multicolumn{2}{c}{\textless Image$_{cpr}$, Image$_{gen}$ \textgreater}& \multicolumn{6}{c}{\textless Image$_{cpr}$, Image$_{unl}$ \textgreater} \\   
    
    \midrule
    \midrule

    \multirow{3}{*}{\textit{Top}}&CM(\%) $\downarrow$ & 98.14 &\textfractionsolidus& 46.80 & 54.01 & 85.79 & 71.93 & 97.11 & 87.45 \\
    
    &$\Delta$CLIP(\%) $\downarrow$ & \textfractionsolidus &\textfractionsolidus& -21.93 & -19.77 & -36.24 & -41.90 & -9.67 & -20.33\\

    &FID $\downarrow$ & 11.18 &\textfractionsolidus& 11.34 & 11.79 & 15.72 & 12.43 & 20.01 & 17.50\\

    \midrule
    \midrule

    \multirow{3}{*}{\textit{Bottom}}&CM(\%) $\downarrow$ & 91.92 & 99.69 & 95.74 & 94.57 & 85.90 & 83.24 & 94.52 & 92.57\\
    
    &$\Delta$CLIP(\%) $\downarrow$ & 5.08 & \textfractionsolidus & -21.56 & -31.18 & -2.08 & -16.17 & -9.76 & -0.63\\
    
    &FID $\downarrow$ & 13.21 & 9.33 & 36.93 & 11.85 & 8.61 & 12.95 & 25.35 & 27.43\\

    \bottomrule[1.5pt]
  \end{tabular}
  \vspace{-3mm}
  \label{benchmark_performance}
\end{table*}

\begin{figure*}[t]
  \centering
  \includegraphics[width=1\linewidth]{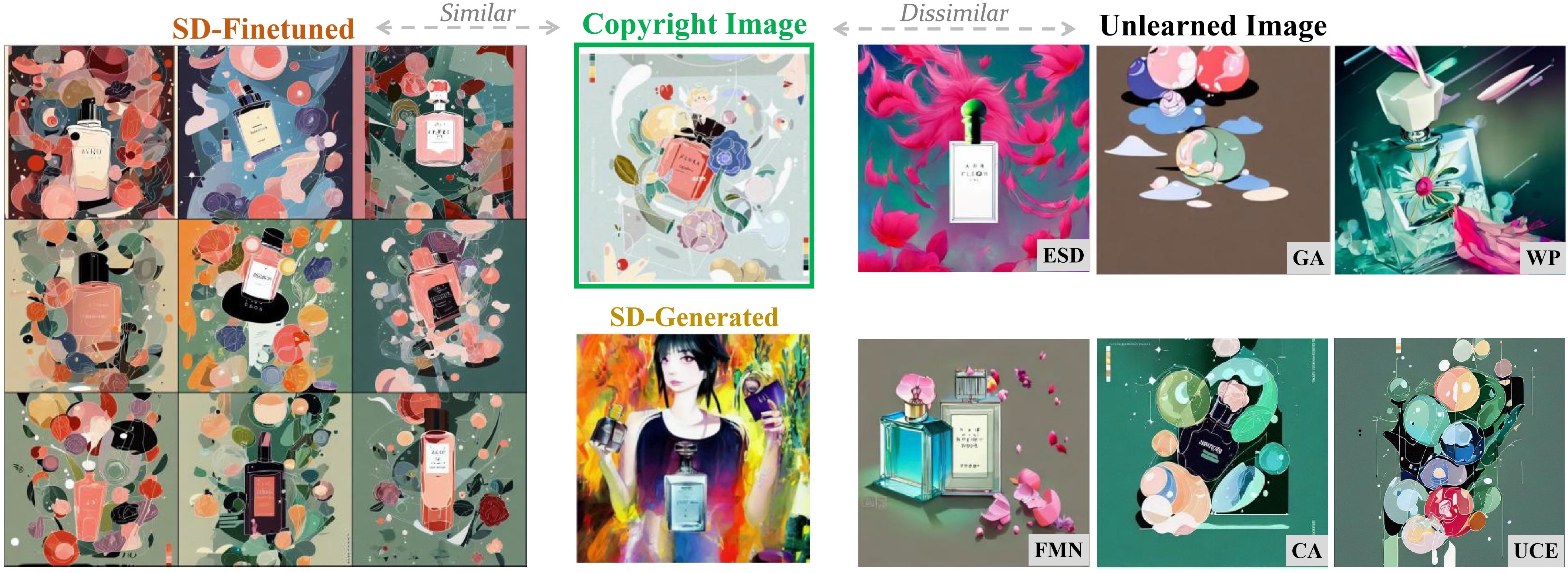}
  \caption{Experimental results of unlearning method on specially finetuned diffusion models. 
  }
  \label{fig:benchmark and illustration}
  \vspace{-5mm}
\end{figure*}


\subsection{Analysis and Interpretation}

\textbf{Interpretation.}
As shown in Fig. \ref{fig:benchmark and illustration}, visualizations of the experimental results for various unlearning algorithms are provided, along with the infringement images generated by the SD-Finetuned.
\textbf{1)} \textit{GA.}
This experiment utilized two foundational models, SD-Finetuned and the stable diffusion \href{https://huggingface.co/stabilityai/stable-diffusion-2-1}{v2.1}. They were respectively employed to evaluate the effectiveness of unlearning algorithms on customized fine-tuned generative models and basic open-source models.
During the unlearning experiments, for each image to be forgotten, the learning rate for GA is set at 5.0e-05, and unlearning training is performed for five epochs.
\textbf{2)}\textit{WP.}
For the SD-Finetuned model, which is based on SD-v1.4 and has been fine-tuned on specific illustration styles, we set the pruning ratios $p_c\%$ and $p_w\%$ to 0.1 and 0.03, respectively. As for SD-v2.1 models, we set the pruning ratios $p_c\%$ and $p_w\%$ to 0.1 and 0.005, respectively. The reason for employing a higher pruning ratio on SD-Finetuned is due to the need for a more significant pruning ratio to forget such artistic styles when the model has been fine-tuned on a limited amount of data. For each image to be forgotten, we performed one epoch of iterative computation and pruning. During the pruning process, the optimizer remained disabled.
\textbf{3) SD-Finetuned.}
During the fine-tuning process of the SD-Finetuned model, we carefully utilized a set of 160 art images along with their corresponding prompts. To more accurately simulate the process of generating infringing artworks, we opted for a two-stage fine-tuning procedure. Implementing this meticulous approach allows us to generate infringing images that exhibit high similarity to the original artworks in artistic features and various image styles.
Specific training details are provided in the appendix.

\textbf{Analysis.}
The bottom of Tab.\ref{benchmark_performance} represents the results on the SD-Finetuned model using CPDM Illustration data, corresponding to the visualizations in Fig.\ref{fig:benchmark and illustration}. A detailed analysis of this part of the experiment reveals the following:
\textbf{1)} The CM metric for images generated directly by the diffusion model is 91.92, while for images generated by SD-Finetuned, the CM metric is as high as 99.69. This indicates that general generative models cannot directly replicate pictures of a specific style, but after fine-tuning, they can imitate the image style very closely.
\textbf{2)} The unlearning effects of the baseline methods \textit{WP} and \textit{GA} are relatively weak. As shown in Fig.\ref{fig:benchmark and illustration}, the results of \textit{WP} and \textit{GA} still contain illustration styles similar to the copyrighted images. Additionally, the \textit{WP} method significantly affects the model's generative capability (refer to the FID changes).
\textbf{3)} Comparing several public unlearning methods, \textit{ESD}, \textit{FMN}, \textit{CA}, and \textit{UCE}, it is found that \textit{FMN} has a better forgetting effect (lower CM in Tab.\ref{benchmark_performance}). Furthermore, it can be observed that the images generated by \textit{FMN} in Fig.\ref{fig:benchmark and illustration} differ significantly in artistic style from the copyrighted images.
\textbf{4)} The forgetting effect of \textit{CA} is weaker compared to other methods (higher CM metric in Tab.\ref{benchmark_performance}). Observing the images generated by \textit{CA} in Fig.\ref{fig:benchmark and illustration}, \textit{CA}'s images feature a representative circular illustration pattern surrounding a perfume bottle, similar to the copyrighted images.
The evaluation metrics and visualization images illustrate the effectiveness of the benchmark.

%% file: sec/3_finalcopy.tex
\subsection{CPDM Metric}

\begin{table}[]
  \centering
  \caption{
    \label{tab:metrics}
    Comprehensive Evaluation Metrics. In this context, \enquote{Image$_{cpr}$}, \enquote{Prompt$_{cpr}$}, \enquote{Model$_{ori}$}, \enquote{Model$_{unl}$}, and \enquote{Model$_{icp}$} represent the copyright image, the prompt associated with the copyright image, the original SD model, the SD model after unlearning, and the Inception model, respectively.
  }
  \setlength{\tabcolsep}{0.45mm}
  \begin{tabular}{c | c  c  c}
    \hline\toprule[1pt]
    \textbf{Input} & \multicolumn{2}{c}{\textbf{Metric}} & \textbf{Description} \\
    
    \midrule
    \multirow{5}{*}{
      \begin{tabular}{c}
        Image$_{cpr}$, Prompt$_{cpr}$ \vspace{1mm}\\
        Image$_{gen}$: 
        Model$_{ori}$ (Prompt$_{cpr}$) \vspace{1mm}\\
        Image$_{unl}$: 
        Model$_{unl}$ (Prompt$_{cpr}$) \vspace{1mm}\\ 
        Model$_{ori}$, Model$_{unl}$, Model$_{icp}$ \vspace{3mm}\\
      \end{tabular}
    } 
    &  \textit{ Potential for Infringement}
    & \textbf{CM} & \textless Image$_{cpr}$, Image$_{gen}$ \textgreater \\
    \cline{2-4}

    & \multirow{3}{*}{\textit{Effectiveness of Unlearning}} 
    & \textbf{CM} & \textless Image$_{cpr}$, Image$_{unl}$ \textgreater \\
    
    & &\textbf{$\Delta$CLIP} & 
      \begin{tabular}{c}
        \textless Image$_{gen}$,  Prompt$_{cpr}$\textgreater\\
        \textless Image$_{unl}$, Prompt$_{cpr}$\textgreater
      \end{tabular}\\
    \cline{2-4}
    
    & \textit{Model Degradation}
    & \textbf{FID} & Model$_{icp}$ (Image$_{unl}$) \\
    
    \bottomrule[1.5pt]
  \end{tabular}
  \vspace{-5mm}
\end{table}


In Tab. \ref{tab:metrics}, the metric consists of CM, $\Delta$CLIP, and FID. As the latter two metrics are apparent in formulation, we focus on illustrating the construction of the CM metric in this section. 
We collaborate with an artist, \textit{Anonymous Artist}, who is currently active in the art industry and is notably distinguished in the field of illustration. This artist offers invaluable insights from a professional standpoint to help assess whether two images constitute plagiarism, considering elements such as brushstrokes, color palettes, lighting effects, and composition. 
However, accurately defining these potentially copyrighted attributes of images is challenging. We initially divide measurements into two perspectives: semantic and stylistic components. The goal is to develop a formula that combines these two components and provides a scalar measure to quantify the similarity between two images.
  

\textbf{Semantic Metric. }
We leverage the CLIP\cite{radford2021learning} model to generate the semantic embedding, and calculate the metrics by: 
\begin{equation}
emb_{ori} = E_{clip}(Image_{ori}), \quad emb_{gen} = E_{clip}(Image_{gen})
\end{equation}
\begin{equation}
M_{\text{sem}} = \frac{1}{n} \sum_{i=1}^{n} (emb_{\text{ori}}[i] - emb_{\text{gen}}[i])^2
\end{equation}
where $Image_{ori}$ and $Image_{gen}$ denote the anchor image and generated image respectively; $E_{clip}$ denotes the CLIP's image encoder; $i$ represents pixel value. In previous studies, cosine similarity has been predominantly used as the evaluation metric. However, in this research, we utilize MSE instead. This choice is motivated by two primary factors: first, the range of MSE is significantly broader than that of cosine similarity, making it easier to observe changes resulting from unlearning; second, adopting MSE aligns better with the subsequent style metrics discussed below.

\textbf{Style Metric. }
Inspired by \cite{gatys2015neural}, we use the activation output of CNN networks to calculate feature correlations via the Gram matrix. We leverage InceptionV3 \cite{szegedy2015rethinking}, following the FID metric.
\begin{equation}
    G^l_{ori} = Gram(E_{incep}(Image_{ori}, l)), \quad
     G^l_{gen} = Gram(E_{incep}(Image_{gen}, l)) 
\end{equation}
\begin{equation}
    D^l = \frac{1}{n} \sum_{i=1}^{n} (G^l_{ori}[i] - G^l_{gen}[i])^2  
\end{equation}
\begin{equation}
    M_{style} = \sum_{i=1}^{n}w^lD^l
\end{equation}
where $E_{incep}(Image,\,l)$ denotes passsing the $Image$ through an Inception network and extracting the feature maps from layer $l$. The Gram matrix is then computed to provide a style representation of the image at layer $l$. Furthermore, the dissimilarity between the original and generated images in each layer is represented by MSE of the Gram matrices in each corresponding layer.
The total style metric, as described above, is determined by weighting factors $w^l$, which represent the contribution of each layer to the overall style metrics. In our work, $n$ is set to $4$, because there are four stages of the InceptionV3 model. The values of parameters $w^l$ needs to be fine-tuned according to the distribution of the images. We provide specific values applicable to our dataset in the supplementary material.
Finally, we denote the total metric as:
\begin{align}
CM &= 1 - Norm(M_{sem} \times M_{style}) ^ 2
\end{align}
Here, $Norm$ represents normalization, which involves clipping data to the range $[a, b]$, then dividing by $(b - a)$, with $a$ and $b$ empirically set to 1 and 50, respectively. The squared term emphasizes the significant changes before and after the unlearning process. The effectiveness of this quantifiable metric will be verified in section \ref{sec:effectiveness}.

%% file: sec/7_conclusion.tex
\section{Conclusion}
The remarkable generation and data fitting capabilities of large models like diffusion models have garnered significant attention, and also raised concerns regarding image copyright. 
Recently, machine unlearning has emerged as a potential solution to mitigate the infringement risk by unlearning the corresponding copyright memory. 
This work introduces a new large-scale dataset and benchmark, coupled with standardized metrics, for evaluating machine unlearning methods, specifically designed to assess the efficacy of machine unlearning methods in safeguarding copyright. 
Notably, this dataset represents the inaugural compilation in this domain that is predicated on diffusion models. The dataset's caliber is substantiated by rigorous human and artist evaluations, ensuring that the metrics resonate with human perception and are aligned with the sensibilities of creators.


%% file: sec/checklist.tex
\clearpage
\newpage
\section*{Checklist}



\begin{enumerate}

\item For all authors...
\begin{enumerate}
  \item Do the main claims made in the abstract and introduction accurately reflect the paper's contributions and scope?
    \answerYes \{See section: The abstract and introduction in the main text, and the scope in Supplementary Materials.\}
  \item Did you describe the limitations of your work?
    \answerYes \{See section: Limitations and Negative Societal Impacts in Supplementary Materials.\}
  \item Did you discuss any potential negative societal impacts of your work?
    \answerYes \{See section: Limitations and Negative Societal Impacts in Supplementary Materials.\}
  \item Have you read the ethics review guidelines and ensured that your paper conforms to them?
    \answerYes{}
\end{enumerate}

\item If you are including theoretical results...
\begin{enumerate}
  \item Did you state the full set of assumptions of all theoretical results?
    \answerYes{\{In the section of defining evaluation metrics to assess image infringement, we provide corresponding theoretical explanations.\}}
	\item Did you include complete proofs of all theoretical results?
    \answerYes{}
\end{enumerate}

\item If you ran experiments (e.g. for benchmarks)...
\begin{enumerate}
  \item Did you include the code, data, and instructions needed to reproduce the main experimental results (either in the supplemental material or as a URL)?
    \answerYes{\{We will open source the corresponding experimental code in \href{https://github.com/rmpku/CPDM}{\textit{github}} and provide detailed descriptions of parameter settings in the experimental setup section to ensure experiment reproducibility.\}}
  \item Did you specify all the training details (e.g., data splits, hyperparameters, how they were chosen)?
    \answerYes \{See section: Experiments setting.\}
	\item Did you report error bars (e.g., with respect to the random seed after running experiments multiple times)?
    \answerYes{}
	\item Did you include the total amount of compute and the type of resources used (e.g., type of GPUs, internal cluster, or cloud provider)?
    \answerYes\{See section: Computational Resources in in Supplementary Materials\}
\end{enumerate}

\item If you are using existing assets (e.g., code, data, models) or curating/releasing new assets...
\begin{enumerate}
  \item If your work uses existing assets, did you cite the creators?
    \answerNo\{The dataset we created does not incorporate any existing publicly available datasets.\}
  \item Did you mention the license of the assets?
    \answerYes\{See section: License in Appendix in Supplementary Materials.\}
  \item Did you include any new assets either in the supplemental material or as a URL?
    \answerYes\{See section: Dataset hosting and maintenance in in Supplementary Materials.\}
  \item Did you discuss whether and how consent was obtained from people whose data you're using/curating?
    \answerYes\{See section: License in in Supplementary Materials.\}
  \item Did you discuss whether the data you are using/curating contains personally identifiable information or offensive content?
    \answerYes\{See section: Limitations and Negative Societal Impacts in Supplementary Materials.\}
\end{enumerate}

\item If you used crowdsourcing or conducted research with human subjects...
\begin{enumerate}
  \item Did you include the full text of instructions given to participants and screenshots, if applicable?
    \answerNo\{Our dataset comprises a specific category of human portraits, sourced from publicly accessible outlets such as Wikipedia, where the images are made available for non-commercial or educational purposes.\}
  \item Did you describe any potential participant risks, with links to Institutional Review Board (IRB) approvals, if applicable?
    \answerNo{}
  \item Did you include the estimated hourly wage paid to participants and the total amount spent on participant compensation?
    \answerYes\{See section: Human Resources in Appendix in Supplementary Materials.\}
\end{enumerate}

\end{enumerate}

%% file: sec/datasheets.tex
\newpage
\clearpage
\section{Datasheets for datasets}




\definecolor{darkblue}{RGB}{100, 100, 200}

\newcommand{\dssectionheader}[1]{%
   \noindent\framebox[\columnwidth]{%
      {\fontfamily{phv}\selectfont \textbf{\textcolor{darkblue}{#1}}}
   }
}

\newcommand{\dsquestion}[1]{%
    {\noindent \fontfamily{phv}\selectfont \textcolor{darkblue}{\textbf{#1}}}
}

\newcommand{\dsquestionex}[2]{%
    {\noindent \fontfamily{phv}\selectfont \textcolor{darkblue}{\textbf{#1} #2}}
}

\newcommand{\dsanswer}[1]{%
   {\noindent #1 \medskip}
}

\begin{singlespace}

\dssectionheader{Motivation}

\dsquestionex{For what purpose was the dataset created?}{Was there a specific task in mind? Was there a specific gap that needed to be filled? Please provide a description.}

\dsanswer{The dataset was created with the purpose of advancing research and development in the field of text-to-image generative models. These models aim to generate realistic images based on textual descriptions, effectively bridging the gap between language and visual content. However, the rapid advancements in text-to-image generation techniques have also raised concerns regarding copyright protection, such as the unauthorized learning of content, artistic creations, and portrait. We aim to develop a dataset and metrics that facilitate the identification of copyright infringement, while enabling a fair comparison of methods for mitigating such infringements.
}

\dsquestion{Who created this dataset (e.g., which team, research group) and on behalf of which entity (e.g., company, institution, organization)?}

\dsanswer{This dataset was collaboratively created by researchers from Peking University, Tsinghua University, and University of California, Berkeley (UCB), as well as researchers from the industry, specifically ByteDance company.
}

\dsquestionex{Who funded the creation of the dataset?}{If there is an associated grant, please provide the name of the grantor and the grant name and number.}

\dsanswer{

No.
}

\dsquestion{Any other comments?}

\dsanswer{No.
}

\bigskip
\dssectionheader{Composition}

\dsquestionex{What do the instances that comprise the dataset represent (e.g., documents, photos, people, countries)?}{ Are there multiple types of instances (e.g., movies, users, and ratings; people and interactions between them; nodes and edges)? Please provide a description.}

\dsanswer{The dataset primarily comprises anchor images, generated images and their corresponding prompts. The anchor images are initially collected as a set of images that potentially contain copyrighted content. These anchor images are then processed using the CLIP-interrogator, which yields prompts associated with each anchor image. Subsequently, the obtained prompts are utilized as inputs for the stable diffusion model, enabling the generation of images by the stable diffusion model.
}

\dsquestion{How many instances are there in total (of each type, if appropriate)?}

\dsanswer{The dataset consists of 2100 anchor images, each accompanied by a corresponding prompt, resulting in a total of 2100 prompts. Using these prompts as input, a total of 18900 images were generated.
}

\dsquestionex{Does the dataset contain all possible instances or is it a sample (not necessarily random) of instances from a larger set?}{ If the dataset is a sample, then what is the larger set? Is the sample representative of the larger set (e.g., geographic coverage)? If so, please describe how this representativeness was validated/verified. If it is not representative of the larger set, please describe why not (e.g., to cover a more diverse range of instances, because instances were withheld or unavailable).}

\dsanswer{ The dataset encompasses the entirety of all possible instances.
}

\dsquestionex{What data does each instance consist of? “Raw” data (e.g., unprocessed text or images) or features?}{In either case, please provide a description.}

\dsanswer{Each instance within the dataset comprises an anchor image, a prompt, and nine corresponding generated images.
}

\dsquestionex{Is there a label or target associated with each instance?}{If so, please provide a description.}

\dsanswer{Each instance represents an original image, along with its corresponding prompt and nine images generated by the stable diffusion model that potentially exhibit copyright infringement.
}

\dsquestionex{Is any information missing from individual instances?}{If so, please provide a description, explaining why this information is missing (e.g., because it was unavailable). This does not include intentionally removed information, but might include, e.g., redacted text.}

\dsanswer{

No.
}

\dsquestionex{Are relationships between individual instances made explicit (e.g., users’ movie ratings, social network links)?}{If so, please describe how these relationships are made explicit.}

\dsanswer{There is no explicit correlation between individual instances.
}

\dsquestionex{Are there recommended data splits (e.g., training, development/validation, testing)?}{If so, please provide a description of these splits, explaining the rationale behind them.}

\dsanswer{No.
}

\dsquestionex{Are there any errors, sources of noise, or redundancies in the dataset?}{If so, please provide a description.}

\dsanswer{No, there are no errors, sources of noise, or redundancies in the dataset.
}

\dsquestionex{Is the dataset self-contained, or does it link to or otherwise rely on external resources (e.g., websites, tweets, other datasets)?}{If it links to or relies on external resources, a) are there guarantees that they will exist, and remain constant, over time; b) are there official archival versions of the complete dataset (i.e., including the external resources as they existed at the time the dataset was created); c) are there any restrictions (e.g., licenses, fees) associated with any of the external resources that might apply to a future user? Please provide descriptions of all external resources and any restrictions associated with them, as well as links or other access points, as appropriate.}

\dsanswer{The dataset is self-contained.
}

\dsquestionex{Does the dataset contain data that might be considered confidential (e.g., data that is protected by legal privilege or by doctor-patient confidentiality, data that includes the content of individuals non-public communications)?}{If so, please provide a description.}

\dsanswer{No, the dataset does not contain data that might be considered confidential, such as information protected by legal privilege, doctor-patient confidentiality, or the content of individuals' non-public communications.
}

\dsquestionex{Does the dataset contain data that, if viewed directly, might be offensive, insulting, threatening, or might otherwise cause anxiety?}{If so, please describe why.}

\dsanswer{No.
}

\dsquestionex{Does the dataset relate to people?}{If not, you may skip the remaining questions in this section.}

\dsanswer{Yes
}

\dsquestionex{Does the dataset identify any subpopulations (e.g., by age, gender)?}{If so, please describe how these subpopulations are identified and provide a description of their respective distributions within the dataset.}

\dsanswer{The dataset primarily comprises four categories: Style, Portrait, Artistic Creation Figure, and Licensed Illustration.
}

\dsquestionex{Is it possible to identify individuals (i.e., one or more natural persons), either directly or indirectly (i.e., in combination with other data) from the dataset?}{If so, please describe how.}

\dsanswer{Yes, portraits are also part of copyright, and therefore, we have included a subset of celebrity portraits in the dataset.
}

\dsquestionex{Does the dataset contain data that might be considered sensitive in any way (e.g., data that reveals racial or ethnic origins, sexual orientations, religious beliefs, political opinions or union memberships, or locations; financial or health data; biometric or genetic data; forms of government identification, such as social security numbers; criminal history)?}{If so, please provide a description.}

\dsanswer{No.
}

\dsquestion{Any other comments?}

\dsanswer{No.
}

\bigskip
\dssectionheader{Collection Process}

\dsquestionex{How was the data associated with each instance acquired?}{Was the data directly observable (e.g., raw text, movie ratings), reported by subjects (e.g., survey responses), or indirectly inferred/derived from other data (e.g., part-of-speech tags, model-based guesses for age or language)? If data was reported by subjects or indirectly inferred/derived from other data, was the data validated/verified? If so, please describe how.}

\dsanswer{The data was directly observable.
}

\dsquestionex{What mechanisms or procedures were used to collect the data (e.g., hardware apparatus or sensor, manual human curation, software program, software API)?}{How were these mechanisms or procedures validated?}

\dsanswer{We propose a pipeline to coordinate CLIP, ChatGPT, and diffusion models to generate a dataset that contains anchor images, corresponding prompts, and images generated by text-to-image models, reflecting the potential abuses of copyright.
Initially, we collect a set of images that potentially contain copyrighted content, which serves as anchor images. Subsequently, these images are fed into the CLIP-interrogator, allowing us to obtain prompts that correspond to each anchor image. Finally, the prompts are used as input for the stable diffusion model, resulting in the generation of images by the stable diffusion model. Through manual comparisons, we assess whether there is evidence of copyright infringement in terms of style and semantics between the anchor images and the generated images. Ultimately, the anchor images, their corresponding prompts, and the images generated by the stable diffusion model constitute the core components of our dataset.
}

\dsquestion{If the dataset is a sample from a larger set, what was the sampling strategy (e.g., deterministic, probabilistic with specific sampling probabilities)?}

\dsanswer{No.
}

\dsquestion{Who was involved in the data collection process (e.g., students, crowdworkers, contractors) and how were they compensated (e.g., how much were crowdworkers paid)?}

\dsanswer{The dataset was primarily curated with contributions from the first three authors listed in the author list.
}

\dsquestionex{Over what timeframe was the data collected? Does this timeframe match the creation timeframe of the data associated with the instances (e.g., recent crawl of old news articles)?}{If not, please describe the timeframe in which the data associated with the instances was created.}

\dsanswer{
The data was collected within the past three months.
}

\dsquestionex{Were any ethical review processes conducted (e.g., by an institutional review board)?}{If so, please provide a description of these review processes, including the outcomes, as well as a link or other access point to any supporting documentation.}

\dsanswer{No.
}

\dsquestionex{Does the dataset relate to people?}{If not, you may skip the remaining questions in this section.}

\dsanswer{Yes.
}

\dsquestion{Did you collect the data from the individuals in question directly, or obtain it via third parties or other sources (e.g., websites)?}

\dsanswer{We obtained portrait information of public figures from Wikipedia.
}

\dsquestionex{Were the individuals in question notified about the data collection?}{If so, please describe (or show with screenshots or other information) how notice was provided, and provide a link or other access point to, or otherwise reproduce, the exact language of the notification itself.}

\dsanswer{Our images are sourced from Wikipedia, where the images are available for non-commercial or educational use.
}

\dsquestionex{Did the individuals in question consent to the collection and use of their data?}{If so, please describe (or show with screenshots or other information) how consent was requested and provided, and provide a link or other access point to, or otherwise reproduce, the exact language to which the individuals consented.}

\dsanswer{Our images are sourced from Wikipedia, where the images are available for non-commercial or educational use.
}

\dsquestionex{If consent was obtained, were the consenting individuals provided with a mechanism to revoke their consent in the future or for certain uses?}{If so, please provide a description, as well as a link or other access point to the mechanism (if appropriate).}

\dsanswer{We will provide our contact information on the release page of the website. In the event of any potential copyright infringement, we will promptly assess the situation, and if found to be valid, we will take immediate action to remove the corresponding data.
}

\dsquestionex{Has an analysis of the potential impact of the dataset and its use on data subjects (e.g., a data protection impact analysis) been conducted?}{If so, please provide a description of this analysis, including the outcomes, as well as a link or other access point to any supporting documentation.}

\dsanswer{We assert that our data is unlikely to cause potential negative impacts.
}

\dsquestion{Any other comments?}

\dsanswer{No.
}

\bigskip
\dssectionheader{Preprocessing/cleaning/labeling}

\dsquestionex{Was any preprocessing/cleaning/labeling of the data done (e.g., discretization or bucketing, tokenization, part-of-speech tagging, SIFT feature extraction, removal of instances, processing of missing values)?}{If so, please provide a description. If not, you may skip the remainder of the questions in this section.}

\dsanswer{

No.
}

\dsquestionex{Was the “raw” data saved in addition to the preprocessed/cleaned/labeled data (e.g., to support unanticipated future uses)?}{If so, please provide a link or other access point to the “raw” data.}

\dsanswer{No.
}

\dsquestionex{Is the software used to preprocess/clean/label the instances available?}{If so, please provide a link or other access point.}

\dsanswer{No.
}

\dsquestion{Any other comments?}

\dsanswer{No.
}

\bigskip
\dssectionheader{Uses}

\dsquestionex{Has the dataset been used for any tasks already?}{If so, please provide a description.}

\dsanswer{This dataset is utilized for evaluating the efficacy of unlearning methods applied to stable diffusion.
}

\dsquestionex{Is there a repository that links to any or all papers or systems that use the dataset?}{If so, please provide a link or other access point.}

\dsanswer{No.
}

\dsquestion{What (other) tasks could the dataset be used for?}

\dsanswer{This dataset can also be utilized to assist in determining whether copyright infringement has occurred.
}

\dsquestionex{Is there anything about the composition of the dataset or the way it was collected and preprocessed/cleaned/labeled that might impact future uses?}{For example, is there anything that a future user might need to know to avoid uses that could result in unfair treatment of individuals or groups (e.g., stereotyping, quality of service issues) or other undesirable harms (e.g., financial harms, legal risks) If so, please provide a description. Is there anything a future user could do to mitigate these undesirable harms?}

\dsanswer{No.
}

\dsquestionex{Are there tasks for which the dataset should not be used?}{If so, please provide a description.}

\dsanswer{No.
}

\dsquestion{Any other comments?}

\dsanswer{No.
}

\bigskip
\dssectionheader{Distribution}

\dsquestionex{Will the dataset be distributed to third parties outside of the entity (e.g., company, institution, organization) on behalf of which the dataset was created?}{If so, please provide a description.}

\dsanswer{No.
}

\dsquestionex{How will the dataset will be distributed (e.g., tarball on website, API, GitHub)}{Does the dataset have a digital object identifier (DOI)?}

\dsanswer{We will release this dataset in github.
}

\dsquestion{When will the dataset be distributed?}

\dsanswer{We plan to release our dataset upon the paper entering the review stage.
}

\dsquestionex{Will the dataset be distributed under a copyright or other intellectual property (IP) license, and/or under applicable terms of use (ToU)?}{If so, please describe this license and/or ToU, and provide a link or other access point to, or otherwise reproduce, any relevant licensing terms or ToU, as well as any fees associated with these restrictions.}

\dsanswer{The dataset is available for non-commercial or educational use.
}

\dsquestionex{Have any third parties imposed IP-based or other restrictions on the data associated with the instances?}{If so, please describe these restrictions, and provide a link or other access point to, or otherwise reproduce, any relevant licensing terms, as well as any fees associated with these restrictions.}

\dsanswer{The dataset is available for non-commercial or educational use.
}

\dsquestionex{Do any export controls or other regulatory restrictions apply to the dataset or to individual instances?}{If so, please describe these restrictions, and provide a link or other access point to, or otherwise reproduce, any supporting documentation.}

\dsanswer{The dataset is available for non-commercial or educational use.
}

\dsquestion{Any other comments?}

\dsanswer{No.
}

\bigskip
\dssectionheader{Maintenance}

\dsquestion{Who will be supporting/hosting/maintaining the dataset?}

\dsanswer{The researcher in this project.
}

\dsquestion{How can the owner/curator/manager of the dataset be contacted (e.g., email address)?}

\dsanswer{We have provided the contact information of the dataset creators on our GitHub website.
}

\dsquestionex{Is there an erratum?}{If so, please provide a link or other access point.}

\dsanswer{No.
}

\dsquestionex{Will the dataset be updated (e.g., to correct labeling errors, add new instances, delete instances)?}{If so, please describe how often, by whom, and how updates will be communicated to users (e.g., mailing list, GitHub)?}

\dsanswer{We will release and update our dataset on GitHub, with a monthly update frequency.
}

\dsquestionex{If the dataset relates to people, are there applicable limits on the retention of the data associated with the instances (e.g., were individuals in question told that their data would be retained for a fixed period of time and then deleted)?}{If so, please describe these limits and explain how they will be enforced.}

\dsanswer{
If our images infringe upon individuals' portrait rights, we will promptly remove the corresponding data after verification.
}

\dsquestionex{Will older versions of the dataset continue to be supported/hosted/maintained?}{If so, please describe how. If not, please describe how its obsolescence will be communicated to users.}

\dsanswer{
With each update, we incorporate the changes based on the original dataset, ensuring that previous versions of the data are preserved.
}

\dsquestionex{If others want to extend/augment/build on/contribute to the dataset, is there a mechanism for them to do so?}{If so, please provide a description. Will these contributions be validated/verified? If so, please describe how. If not, why not? Is there a process for communicating/distributing these contributions to other users? If so, please provide a description.}

\dsanswer{On the dataset's release page, we have provided corresponding links with the aim of encouraging collaborative expansion of the dataset and fostering the protection of copyright information.
}

\dsquestion{Any other comments?}

\dsanswer{No.
}

\end{singlespace}